\documentclass[letterpaper]{article} 
\usepackage{aaai2026}  
\usepackage{times}  
\usepackage{helvet}  
\usepackage{courier}  
\usepackage[hyphens]{url}  
\usepackage{graphicx} 
\usepackage{natbib}  
\usepackage{caption} 
\usepackage{algorithm}          
\usepackage[noend]{algpseudocode}   

\usepackage{enumitem}  

\usepackage{amsmath}   
\usepackage{amssymb}   
\usepackage{bm}        
\usepackage{booktabs}  
\usepackage{xcolor}
\providecommand{\Comment}[1]{\hfill$\triangleright$~#1}

\usepackage{newfloat}
\usepackage{listings}
\DeclareCaptionStyle{ruled}{labelfont=normalfont,labelsep=colon,strut=off} 
\floatstyle{ruled}
\newfloat{listing}{tb}{lst}{}
\floatname{listing}{Listing}
\newcommand{\cmark}{\checkmark}
\newcommand{\pcmark}{\resizebox{0.8em}{!}{\large($\checkmark$)}}
\newcommand{\xmark}{---}

\title{Learning to Collaborate: An Orchestrated-Decentralized Framework for Peer-to-Peer LLM Federation}
\author{
    Inderjeet Singh$^{\dagger}$,
    Eleonore Vissol-Gaudin,
    Andikan Otung,
    Motoyoshi Sekiya
}
\affiliations{
    Fujitsu Research of Europe\\
    Slough, United Kingdom\\
    \{inderjeet.singh, eleonore.gaudin, andikan.otung, motoyoshi.sekiya\}@fujitsu.com
}

\begin{document}

\maketitle
\let\thefootnote\relax
\footnotetext{\hspace*{-\parindent}$^{\dagger}$Corresponding author.}
\begin{abstract}
Fine-tuning Large Language Models (LLMs) for specialized domains is constrained by a fundamental challenge: the need for diverse, cross-organizational data conflicts with the principles of data privacy and sovereignty. While Federated Learning (FL) provides a framework for collaboration without raw data exchange, its classic centralized form introduces a single point of failure and remains vulnerable to model inversion attacks. Decentralized FL (DFL) mitigates this risk by removing the central aggregator but typically relies on inefficient, random peer-to-peer (P2P) pairings, forming a collaboration graph that is blind to agent heterogeneity and risks negative transfer. This paper introduces KNEXA-FL, a novel framework for \emph{orchestrated decentralization} that resolves this trade-off. KNEXA-FL employs a non-aggregating Central Profiler/Matchmaker (CPM) that formulates P2P collaboration as a contextual bandit problem, using a LinUCB algorithm on abstract agent profiles to learn an optimal matchmaking policy. It orchestrates direct knowledge exchange between heterogeneous, PEFT-based LLM agents via secure distillation, without ever accessing the models themselves. Our comprehensive experiments on a challenging code generation task show that KNEXA-FL yields substantial gains, improving Pass@1 by $\approx50\%$ relative to random P2P collaboration. Critically, our orchestrated approach demonstrates stable convergence, in stark contrast to a powerful centralized distillation baseline which suffers from catastrophic performance collapse. Our work establishes adaptive, learning-based orchestration as a foundational principle for building robust and effective decentralized AI ecosystems.
\end{abstract}

%
\begin{links}\label{code-link}
     \link{Code}{https://github.com/FujitsuResearch/knexa-fl}
\end{links}

\begin{figure*}[t]
    \centering
    \includegraphics[width=.97\linewidth]{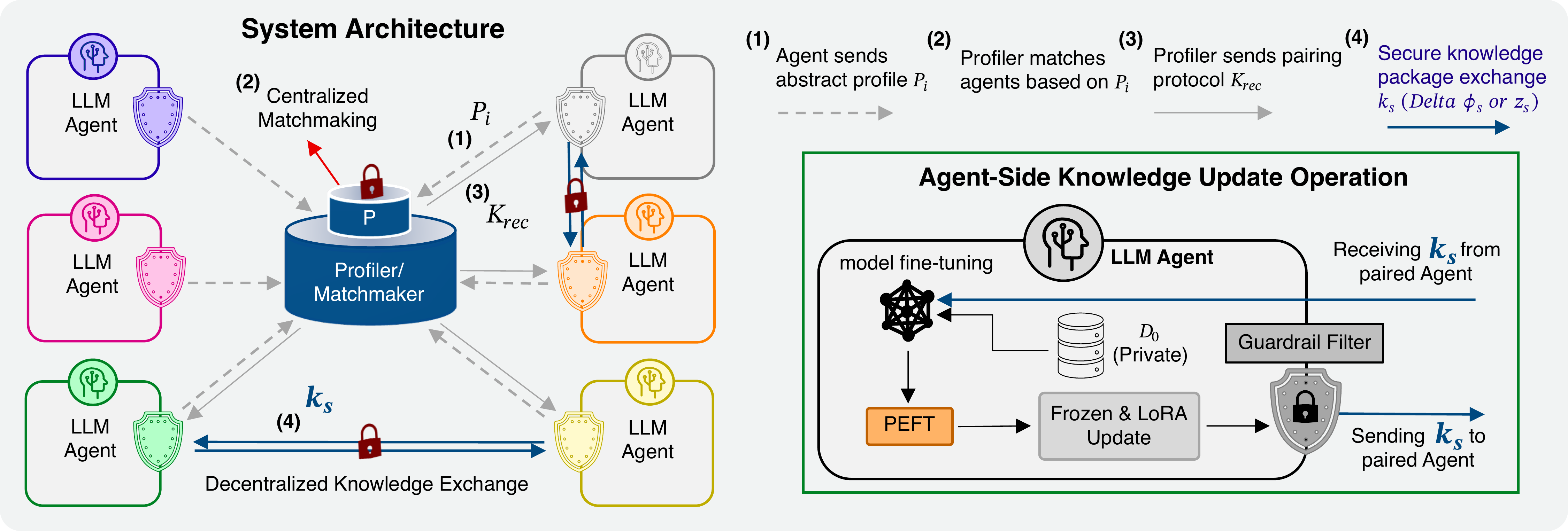}
    \caption{The KNEXA-FL architecture for orchestrated decentralization. Agents send abstract profiles ($\mathbf{p}_i$) to a non-aggregating Central Profiler/Matchmaker (CPM), which provides pairing directives ($K_{rec}$). Paired agents then conduct direct, secure P2P knowledge exchange ($k_s$). The agent-side view (right) details local PEFT fine-tuning and the privacy-enforcing Guardrail Filter.}
    \label{fig:system_architecture}
\end{figure*}

\section{Introduction}
\label{sec:introduction}

Adapting Large Language Models (LLMs) to specialized domains presents a fundamental dilemma: the need for diverse, cross-organizational data clashes with inviolable principles of data sovereignty, privacy, and security. While Federated Learning (FL)~\cite{McMahan2017FedAvg} offers a paradigm for collaborative training without raw data exchange, its canonical server-centric form introduces a trusted central aggregator. This entity becomes a single point of failure and a privileged attack surface for sophisticated model inversion attacks that can reconstruct sensitive training data~\cite{geiping2020inverting,zhao2024loki}.

Decentralized FL (DFL) addresses this by eschewing a central server in favor of direct peer-to-peer (P2P) communication~\cite{Roy2019BrainTorrent,Itahara2023P2PKD}. However, in removing the central aggregator, DFL architectures typically regress to simplistic interaction strategies, such as random or static peer pairings. This random-walk approach to collaboration is statistically inefficient, blind to agent heterogeneity, and risks \emph{negative transfer}, where poorly matched peers degrade each other's performance. It fails to strategically harness the network's latent collective intelligence.

This paper argues that the prevailing dichotomy between a vulnerable central aggregator and inefficient random P2P interaction is too limited. We introduce \textsc{KNEXA-FL}, a novel hybrid framework that enables \emph{orchestrated decentralization}. It features a non-aggregating \textbf{Central Profiler/Matchmaker (CPM)} whose sole purpose is to learn an optimal matchmaking policy for a federation of heterogeneous, autonomous LLM agents. The CPM operates on abstract, privacy-preserving agent profiles to solve a contextual bandit problem, intelligently routing knowledge exchange without ever accessing or storing the models themselves. The actual knowledge transfer, via prediction distillation compatible with parameter-efficient fine-tuning (PEFT), occurs directly and securely between the matched peers.

Our contributions are threefold:
\begin{itemize}
    \item We designed and propose \textsc{KNEXA-FL}, a hybrid, centrally orchestrated - decentralized learning architecture for collaborative LLM fine-tuning. It resolves the matchmaking inefficiency of DFL without re-introducing the security vulnerabilities of a central model aggregator, thus offering a more robust and effective collaboration paradigm.
    \item We are the first to formulate and solve the P2P LLM collaboration problem as a contextual bandit. Our CPM employs a LinUCB algorithm to learn a pairing policy from interaction rewards, dynamically optimizing the P2P knowledge graph to maximize network utility.
    \item We provide comprehensive empirical validation on a challenging, heterogeneous code generation task. Our experiments show that \textsc{KNEXA-FL} substantially outperforms isolated and random P2P baselines, and, critically, demonstrates stable convergence where a state-of-the-art centralized knowledge distillation approach suffers from catastrophic collapse.
\end{itemize}

\section{Related Work}
\label{sec:related_work}

\textsc{KNEXA-FL} integrates ideas from four strands of literature: \emph{(i)} decentralized / peer-to-peer federated learning, \emph{(ii)} parameter-efficient LLM federation, \emph{(iii)} security‐aware governance in collaborative AI, and \emph{(iv)} orchestration, data spaces, and multi-agent systems. We briefly position our contributions in each area.

\subsection{Federated and Decentralized Learning}
Federated Learning (FL) enables collaborative model training by exchanging parameter updates rather than raw data~\cite{McMahan2017FedAvg}. Canonical server-centric FL, however, struggles with statistical heterogeneity (non-IID data) and system heterogeneity (diverse model architectures). While methods like FedProx~\cite{Li2020FedProx} and SCAFFOLD~\cite{Karimireddy2020SCAFFOLD} address statistical drift, they maintain a central aggregator. Similarly, server-mediated knowledge distillation (e.g., FedMD~\cite{Li2019FedMD}) handles architectural diversity but preserves the central server as a single point of failure and a potential privacy bottleneck.

Decentralized FL (DFL) obviates the server, typically relying on gossip-based protocols where agents average parameters with random neighbors~\cite{hegedHus2021decentralized,Roy2019BrainTorrent}. Such methods generally assume model homogeneity and do not perform intelligent peer selection. More recent P2P frameworks have introduced greater flexibility through knowledge distillation~\cite{Itahara2023P2PKD}, sub-network exchange~\cite{Belal2023P2PFedMask}, or topology-aware aggregation~\cite{ryabinin2021moshpit}. Despite these advances, the fundamental problem of \emph{matchmaking} remains underdeveloped, often relying on random or handcrafted heuristics. 

This is the critical gap our work addresses. In sharp contrast to prior art, our \textbf{CPM} formulates peer selection as a contextual bandit problem, learning to construct a high-utility interaction graph dynamically. Existing learning-based schedulers like Oort~\cite{Lai2021Oort} and FedBalancer~\cite{shin2022fedbalancer} optimize \emph{server-client} selection for system metrics like throughput, a fundamentally different problem. The most relevant prior work, IPLS~\cite{pappas2021ipls}, performs a \emph{single, static peer grouping} at the outset based on model similarity. \textsc{KNEXA-FL} is the first framework to employ online learning to \emph{continually} orchestrate the P2P collaboration graph, adapting to evolving agent knowledge while remaining fully non-aggregating.

\subsection{Parameter-Efficient LLM Federation}
The immense scale of LLMs has catalyzed research into federated fine-tuning using PEFT techniques. Centralized frameworks like FATE-LLM~\cite{Liu2023FATELLM} and FedLoRA variants~\cite{Yang2024FedPepTAO} federate PEFT modules (e.g., LoRA adapters) but rely on a central server for aggregation, which can lead to destructive parameter interference. Decentralized approaches such as FedSKD~\cite{Weng2025FedSKD} and KD-PDFL~\cite{Jeong2023KDPDFL} use P2P knowledge distillation, but as noted, they lack an intelligent matchmaking mechanism. \textsc{KNEXA-FL} advances this line of work by uniquely combining support for heterogeneous LLM backbones and PEFT methods with CPM-guided P2P transfers, eliminating the risks of centralized aggregation while maximizing the efficacy of knowledge exchange.

\subsection{Security and Governance in Collaborative AI}
Security for FL and deployed ML systems covers AML risk analysis and robustness of perception/representation components~\cite{bitton2023evaluating,singh2024advancing}, Byzantine-robust and cryptographic aggregation~\cite{pillutla2022robust,Bonawitz2017SecureAgg}, and decentralized trust via reputation or blockchain mechanisms~\cite{Kang2019ReputationFL,Katevas2020PoliFL}. However, these approaches do not solve the problem of efficient partner selection. \textsc{KNEXA-FL} integrates governance directly into the learning process: the CPM's bandit model naturally learns to down-weight low-utility or malicious peers by observing their associated rewards. This is complemented by agent-side \emph{Guardrail Filters} that enforce local data sovereignty policies. This learning-based approach to governance, where trust is an emergent property of observed utility, distinguishes our work from systems with static, rule-based policies.

\subsection{Orchestration, Data Spaces, and Multi-Agent Systems}
The vision of International Data Spaces~\cite{Otto2022DesignGAIAX} emphasizes data sovereignty through policy-governed exchange. \textsc{KNEXA-FL} provides a concrete mechanism to realize a \emph{knowledge data space}, where the exchanged assets are not raw data but abstract knowledge representations (distilled predictions: logits or decoded text), governed by policies enforced at both the agent level (Guardrails) and the system level (CPM orchestration).

From a mathematical and systems perspective, our framework can be viewed through the lens of Multi-Agent Systems (MAS). The problem of finding effective collaborators is a classic challenge in MAS, often addressed via coalition formation~\cite{georgio2025coral} or task allocation. However, much of this work does not contend with the unique constraints of federated learning, namely extreme statistical heterogeneity and strict privacy requirements. Incentive-driven FL~\cite{zhan2021survey} uses economic mechanisms to encourage participation but does not typically solve the problem of \emph{who} should collaborate with whom for maximal learning efficacy. \textsc{KNEXA-FL} recasts the MAS coordination problem for federated LLMs as a contextual bandit problem. This is a significant departure from prior art: instead of relying on predefined rules or complex negotiation protocols, we leverage online learning to solve the coordination problem directly, optimizing for empirical performance in a dynamic, privacy-critical environment.

\begin{table*}[t]
\centering
\small
\renewcommand{\arraystretch}{1.15} 
\setlength{\tabcolsep}{4.5pt} 
\begin{tabular}{@{}l ccc cc cc@{}}
\toprule
& \multicolumn{3}{c}{\textbf{Core Architecture}} & \multicolumn{2}{c}{\textbf{Key Capabilities}} & \multicolumn{2}{c}{\textbf{Focus Area}} \\
\cmidrule(r){2-4} \cmidrule(lr){5-6} \cmidrule(l){7-8}
\textbf{Method} & 
\textbf{\begin{tabular}[c]{@{}c@{}}Decentral.\\ Arch.\end{tabular}} & 
\textbf{\begin{tabular}[c]{@{}c@{}}P2P\\ Exchange\end{tabular}} & 
\textbf{\begin{tabular}[c]{@{}c@{}}Adaptive\\ Matchmaking\end{tabular}} & 
\textbf{\begin{tabular}[c]{@{}c@{}}Heterog.\\ Support\end{tabular}} & 
\textbf{Governance} & 
\textbf{Theory} & 
\textbf{\begin{tabular}[c]{@{}c@{}}LLM-\\ Native\end{tabular}} \\ \midrule
\textbf{KNEXA-FL (ours)} & \cmark & \cmark & \cmark (\textit{Learned}) & \cmark & \cmark & \cmark & \cmark \\ \midrule
\textit{Centralized Schedulers} \\
Oort~\cite{Lai2021Oort} & \xmark & \xmark & \cmark (\textit{Learned}) & \xmark & \pcmark & \xmark & \xmark \\
\midrule
\textit{Decentralized Systems} \\
GossipLearn~\cite{hegedHus2019gossip} & \cmark & \cmark & \xmark & \xmark & \pcmark & \cmark & \xmark \\
IPLS~\cite{pappas2021ipls} & \pcmark & \cmark & \pcmark (\textit{Static}) & \pcmark & \xmark & \xmark & \xmark \\
SparSFA~\cite{Wang2023SparSFA} & \cmark & \cmark & \xmark & \xmark & \cmark & \xmark & \xmark \\
KD-PDFL~\cite{Jeong2023KDPDFL} & \cmark & \cmark & \xmark & \cmark & \xmark & \xmark & \xmark \\
FedSKD~\cite{Weng2025FedSKD} & \cmark & \cmark & \xmark & \cmark & \xmark & \xmark & \pcmark \\
\midrule
\textit{Centralized Baselines} \\
FedMD~\cite{Li2019FedMD} & \xmark & \pcmark & \xmark & \cmark & \xmark & \xmark & \pcmark \\
\bottomrule
\end{tabular}
\caption{Comparison with representative FL systems. \textsc{KNEXA-FL} is the first to integrate learned adaptive matchmaking into a decentralized, heterogeneous, LLM-native P2P framework. Legend: \cmark = fully; \pcmark = partial; \xmark = not addressed.}
\label{tab:novelty_matrix}
\end{table*}

\section{The KNEXA-FL Framework}
\label{sec:framework}

The KNEXA-FL framework facilitates the collaborative enhancement of heterogeneous Large Language Models (LLMs) by enabling direct, orchestrated knowledge exchange among autonomous agents. It operates as a decentralized knowledge data space, circumventing data or model centralization. Interactions are guided by a CPM that analyzes dynamic, abstract agent profiles to recommend optimal P2P pairings and knowledge exchange protocols. This combination of adaptive P2P mechanisms with a learning-based central orchestrator addresses the dual challenges of LLM heterogeneity and effective, privacy-preserving knowledge sharing.

\subsection{Problem Setting and System Architecture}
\label{subsec:problem_and_arch}
We consider a dynamic set of $N_t$ LLM agents $\mathcal{A}_t=\{a_1,\dots,a_{N_t}\}$ at round $t$. Each agent $a_i$ holds a frozen base model $W_0$ and a trainable PEFT module $\bm{\phi}_i$, primarily LoRA~\cite{Hu2021LoRA}, on its private non-IID dataset $D_i$. The collective goal is a constrained multi-objective optimisation:
\begin{align}
\min_{\{\bm{\phi}_i\}} \sum_{i=1}^{N_t} w_i \mathbb{E}_{(x,y)\sim D_i} \!\bigl[\ell(M_i(x),y)\bigr] 
\label{eq:global_obj}
\end{align}
subject to a set of constraints including: privacy leakage, communication, and computation budgets.

The system architecture, illustrated in Figure~\ref{fig:system_architecture}, comprises three logical components. \textbf{(1) LLM Agents} ($\mathcal{A}$) are autonomous entities, each operating via a Local Gateway that enforces local policies and includes a Guardrail Filter to prevent sensitive data disclosure. \textbf{(2) The CPM  ($\mathcal{P}$)} is a trusted, non-aggregating entity that receives abstract profiles $\mathbf{p}_i$ to orchestrate P2P interactions without accessing raw data or models. \textbf{(3) Secure P2P Communication Channels} are established directly between paired agents for ephemeral, encrypted knowledge exchange. This architecture establishes a managed data space where agents are sovereign knowledge providers, and $\mathcal{P}$ is the intelligent interaction broker.

\subsection{P2P Collaboration Protocol}
\label{subsec:p2p_protocol}
The KNEXA-FL protocol unfolds in iterative rounds of local training, profiling, matchmaking, and P2P knowledge exchange.

\paragraph{Agent-Side Operations.}
Each agent $a_i$ first performs local PEFT updates by fine-tuning its module $\bm{\phi}_i$ on its private data $D_i$: $\bm{\phi}_i \leftarrow \bm{\phi}_i - \eta_L \nabla_{\bm{\phi}_i} \mathcal{L}_i(W_0, \bm{\phi}_i; D_i)$. Subsequently, it prepares a knowledge package $\kappa_i$ for sharing. For Adaptive Knowledge Distillation (AKD), this package contains \emph{teacher predictions} produced on a shared, privacy-vetted transfer set $\mathcal{X}_u$. These predictions may be represented as logits $\mathbf{z}_i(\mathbf{x})$ or as decoded text sequences $y_i(\mathbf{x})$; in our implementation we use the latter. All payloads are vetted by the agent's Guardrail Filter before packaging.

\paragraph{Adaptive Knowledge Distillation (AKD).}
AKD, our core exchange mechanism, uses \textbf{text-based distillation} for robustness to architectural and tokenizer heterogeneity. A teacher agent $a_j$ generates text predictions $y_j(x)$ for each prompt $x \in \mathcal{X}_u$, which are sent to student $a_i$. The student $a_i$ re-encodes the teacher's text $y_j(x)$ with its own tokenizer to create a "soft" target token sequence $\tilde{y}_j(x)$. It then minimizes a standard token-level cross-entropy loss, forcing its own output distribution $p_i(\cdot \mid x)$ to match the teacher's sequence:
\begin{equation}
\label{eq:kd_loss}
\resizebox{0.9\columnwidth}{!}{$
\mathcal{L}^{\textsc{kd}}_{\text{total},i}
= (1-\alpha_{\textsc{kd}})\,\mathcal{L}_i(D_i)
\;+\;
\alpha_{\textsc{kd}}\,\mathbb{E}_{x\in \mathcal{X}_u}
\Big[ \mathcal{L}_{\text{CE}}\big(\tilde{y}_j(x),\, p_i(\cdot \mid x)\big) \Big],$}
\end{equation}
where $\mathcal{L}_{\text{CE}}$ is the token-level cross-entropy loss w.r.t. the re-encoded teacher tokens $\tilde{y}_j(x)$. This text-level approach makes the objective well-defined for any pair, sidestepping all tokenizer mismatch issues.

\begin{table*}[t]
\centering
\resizebox{\textwidth}{!}{%
\begin{tabular}{@{}lcccc@{}}
\toprule
\textbf{Mechanism} & \textbf{Comm.\ Cost} & \textbf{Comp.\ Overhead} & \textbf{Heterogeneity Tol.} & \textbf{Info Specificity / Privacy Risk} \\
\midrule
\textbf{AKD (Teacher predictions)} & Med-High(logits)/Low(text) & Med-High (student train) & \textbf{High} & Medium \\
PEFT Module ($\Delta\bm{\phi}$) & Low (e.g., MBs) & Low (Merge) & Low-Med (Needs $\mathbf{T}_{ij}$) & High (Specific param. changes) \\
\bottomrule
\end{tabular}%
}
\caption{Comparison of Knowledge Exchange Mechanisms. AKD is the primary mechanism in KNEXA-FL due to its high heterogeneity tolerance and more favorable privacy-utility trade-off.}
\label{tab:ke_comparison_detail}
\end{table*}

\subsection{The Central Profiler/Matchmaker (CPM)}
\label{subsec:profiler_matchmaker}
The CPM is the learning-based orchestrator. It intelligently pairs agents by solving a contextual combinatorial bandit problem, moving beyond random or heuristic matchmaking.

\paragraph{Agent Profiles and Contextual Bandit.}
Each agent $a_i$ sends a privacy-preserving profile vector $\mathbf{p}_i \in \mathbb{R}^{d_p}$ to the CPM. This profile concatenates \textbf{static features} (e.g., LLM family, PEFT config), \textbf{dynamic features} (e.g., task performance, perplexity, privacy-preserving embeddings of local data distributions), and \textbf{historical/trust features} (e.g., success rates of past P2P interactions, CPM-maintained trust score). For a potential pair $(a_i, a_j)$, the CPM forms a context vector $\mathbf{x}_{ij}^{(t)} = \varphi(\mathbf{p}_i^{(t)}, \mathbf{p}_j^{(t)}, S_{net}^{(t)})$, capturing their compatibility and the global network state.

\paragraph{LinUCB-based Matchmaking.}
We employ LinUCB~\cite{Li2010LinUCB} to select a set of $K_p$ disjoint pairs per round that maximize expected utility. The CPM models the expected reward of a pairing as $\hat{r}_{ij} = \hat{\bm{\theta}}^{\top}\mathbf{x}_{ij}$ and selects pairs based on an upper confidence bound (UCB) score to balance exploitation and exploration. The reward signal $r_{ij}^{(t)}$ provided by the receiving agent $a_i$ after an exchange with $a_j$ is a scalar value reflecting the utility of the interaction:
\begin{equation}
r_{ij}^{(t)} \;=\; \gamma\, (\mathcal{L}_i^{\text{pre}}-\mathcal{L}_i^{\text{post}}) \;-\; \delta\, \text{KB}_{ij}^{(t)},
\label{eq:reward_def}
\end{equation}
where the first term is the local loss reduction and the second penalizes communication cost ($\text{KB}_{ij}^{(t)}$). The CPM updates its bandit parameters $(\mathbf{A}, \mathbf{b})$ based on observed rewards, progressively learning the optimal matchmaking policy. The detailed matchmaking logic is presented in the Appendix.

\subsection{Overall Protocol and Complexity}
\label{subsec:overall_protocol}
The complete KNEXA-FL protocol is specified in Algorithm~\ref{alg:knexa_protocol}. It formalizes the asynchronous, multi-phase loop involving parallel agent computation, centralized learning-based matchmaking, decentralized knowledge exchange, and the feedback mechanism that drives adaptation.

\begin{algorithm}[tb]
\caption{The KNEXA-FL Protocol}
\label{alg:knexa_protocol}
\begin{algorithmic}[1]
\State \textbf{Initialize:} Profiler $\mathcal{P}$ with LinUCB state $(\mathbf{A} \gets \mathbf{I}_{d_p}, \mathbf{b} \gets \mathbf{0})$.
\Procedure{AgentUpdate}{$a_i, D_i, \bm{\phi}_i$}
    \State $\bm{\phi}_i \gets \bm{\phi}_i - \eta_L \nabla_{\bm{\phi}_i} \mathcal{L}_i(W_0, \bm{\phi}_i; D_i)$ \Comment{Local PEFT update}
    \State $\mathbf{p}_i \gets \text{GenProfile}(\bm{\phi}_i, D_i)$ \Comment{Generate abstract profile}
    \State \textbf{return} $\mathbf{p}_i$
\EndProcedure

\Statex
\For{each communication round $t=1,2,\dots,T$}
    \Statex \textit{// Phase 1: Asynchronous Profiling}
    \ForAll{agent $a_i\in\mathcal{A}$ \textbf{in parallel}}
        \State $\mathbf{p}_i^{(t)} \gets \text{AgentUpdate}(a_i, D_i, \bm{\phi}_i^{(t-1)})$
        \State Send profile $\mathbf{p}_i^{(t)}$ to $\mathcal{P}$.
    \EndFor
    
    \Statex \textit{// Phase 2: Centralized Matchmaking}
    \State $\mathcal{P}$ computes $\hat{\bm{\theta}} \gets \mathbf{A}^{-1}\mathbf{b}$ from its current state.
    \State $\mathcal{P}$ forms pairs $\mathcal{E}_t = \{(a_s, a_r, \mathbf{x}_{sr})\}$ by selecting $K_p$ disjoint pairs that greedily maximize the LinUCB score: $\hat{\bm{\theta}}^{\top}\mathbf{x} + \beta \sqrt{\mathbf{x}^{\top}\mathbf{A}^{-1}\mathbf{x}}$.
    \State $\mathcal{P}$ dispatches matchmaking directives to agents in $\mathcal{E}_t$.

    \Statex \textit{// Phase 3 \& 4: P2P Exchange and Policy Update}
    \ForAll{pair $(a_s, a_r, \mathbf{x}_{sr}) \in \mathcal{E}_t$ \textbf{in parallel}}
        \State $a_r$ receives knowledge package $\kappa_s$ from $a_s$ and integrates it via AKD (Eq.~\ref{eq:kd_loss}).
        \State $a_r$ computes reward $r_{sr}^{(t)}$ via Eq.~\ref{eq:reward_def}.
        \State $a_r$ reports feedback $(\mathbf{x}_{sr}, r_{sr}^{(t)})$ to $\mathcal{P}$.
        \State \textit{On feedback receipt, $\mathcal{P}$ updates its model:}
        \State $\mathbf{A} \leftarrow \mathbf{A} + \mathbf{x}_{sr}(\mathbf{x}_{sr})^\top$; $\mathbf{b} \leftarrow \mathbf{b} + r_{sr}^{(t)}\mathbf{x}_{sr}$.
    \EndFor
\EndFor
\end{algorithmic}
\end{algorithm}

\paragraph{Complexity.}
The communication cost of a P2P exchange is dominated by the AKD payload. When sharing logits, this is $\mathcal{O}(|\mathcal{X}_u|\cdot|V|)$ bytes (e.g., FP16), which is about 62.5 MB for typical values. When sharing decoded text, the cost is $\mathcal{O}(|\mathcal{X}_u|\cdot L_{\text{avg}})$ tokens, which is substantially smaller in practice. The CPM's computational overhead is modest. Each LinUCB feedback update is $\mathcal{O}(d_p^2)$, where $d_p$ is the profile dimensionality. The matchmaking step, if naively enumerating all $\binom{N_t}{2}$ pairs, would be $\mathcal{O}(N_t^2 d_p)$. However, by pre-filtering candidates using efficient approximate nearest neighbor search on profile embeddings, the practical complexity is reduced to a tractable $\mathcal{O}(N_t k d_p)$ for a small neighborhood size $k \ll N_t$. Our implementation with 20 agents completes a full round in under 16 minutes on a cluster of eight A100 GPUs.

\subsection{Security, Privacy, and Theoretical Insight}
\label{subsec:security_and_theory}
KNEXA-FL's security is enhanced by several design principles. \textbf{Data Minimization} is achieved by exchanging only teacher predictions (logits or decoded text), never raw data. \textbf{Secure Communication} (e.g., mTLS with E2E payload encryption) ensures the CPM cannot decrypt knowledge packages. The \textbf{Non-Aggregating CPM} design mitigates central-point-of-failure risks. \textbf{Controlled Influence} is managed via the bandit, which learns to deprioritize malicious or low-quality peers, and through \textbf{Local Gateway Guardrail Filters} that scan outgoing knowledge packages for sensitive information. Future work can enhance \textbf{Verifiability} with DP-noise on logits or ZKPs for profile attestations.

Theoretically, the framework's convergence can be analyzed by viewing the CPM as inducing a dynamic collaboration graph $\mathcal{G}_t$. The LinUCB regret bounds~\cite{Li2010LinUCB} ensure the CPM efficiently learns to form graphs with high-utility edges (i.e., positive expected rewards). Drawing from decentralized consensus literature~\cite{Boyd2006RandomizedGossip}, the system's performance improvement is linked to the spectral properties (e.g., Fiedler value $\lambda_2$) of $\mathcal{G}_t$. As the CPM learns and adds more beneficial edges, $\mathbb{E}[\lambda_2(\mathcal{G}_t)]$ is non-decreasing, suggesting a trajectory towards monotonic performance improvement across the federation. We stress that this is an intuitive linkage to provide insight, not a formal convergence proof for the combined system.

\begin{figure*}[t]
    \centering
    \includegraphics[width=1.0\linewidth]{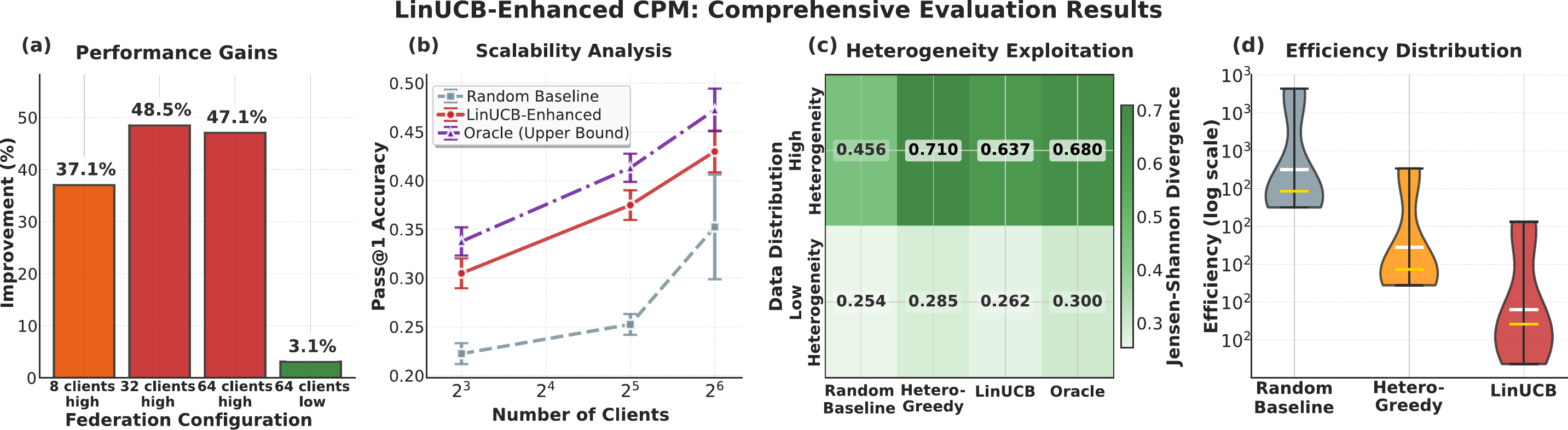} 
    \caption{LinUCB-enhanced CPM comprehensive evaluation in a synthetic federation.
  (a) {Performance Gains:} The relative improvement over random pairing peaks at 48.5\% for 32 clients in a high-heterogeneity setting and remains substantial (37.1\%) even for 8 clients. Gains are modest (3.1\%) in low-heterogeneity scenarios, confirming the CPM's primary value is in exploiting diversity.
  (b) {Scalability Analysis:} Pass@1 accuracy demonstrates our LinUCB approach consistently outperforms the random baseline across all federation sizes (8 to 64 clients) and robustly tracks towards the oracle upper bound.
  (c) {Heterogeneity Exploitation:} The Jensen-Shannon (JS) divergence of selected pairs reveals the CPM's learned strategy. While a naive \texttt{Hetero-Greedy} baseline maximizes JS divergence (0.710), our approach learns a superior trade-off, maintaining high divergence ($\approx$0.64) while selecting for synergistic compatibility.
  (d) {Efficiency Distribution:} The computational efficiency, measured in rewards processed per second, remains highly practical, confirming that the CPM's intelligence does not introduce prohibitive overhead.
  Error bars and shaded regions represent 95\% confidence intervals over five independent runs.}
    \label{fig:synthetic_ablation}
\end{figure*}

\section{Experiments}
\label{sec:experiments}

We conduct a comprehensive empirical study to validate \textsc{KNEXA-FL}, designed to answer three primary research questions:

\begin{description}[leftmargin=0.6cm]
\item[\textbf{RQ1:} \emph{Overall Performance}] Does profiler-guided P2P collaboration in \textsc{KNEXA-FL} achieve superior task performance compared to isolated training, unguided P2P collaboration, and a conventional centralized knowledge distillation baseline?
\item[\textbf{RQ2:} \emph{Matchmaking Efficacy}] To what extent are the performance gains attributable to the intelligent matchmaking of the Central Profiler/Matchmaker over heuristic or random pairing strategies?
\item[\textbf{RQ3:} \emph{Robustness \& Scalability}] How does \textsc{KNEXA-FL}'s performance scale with the size of the federation and the degree of model and data heterogeneity among clients?
\end{description}

\subsection{Experimental Setup}
\label{subsec:exp_setup}

\paragraph{Datasets and Tasks.} We focus on code generation, a challenging domain that demands complex reasoning and syntactic precision. We construct our primary dataset by merging the \textbf{HumanEval}~\cite{chen2021evaluating} and \textbf{MBPP}~\cite{austin2021program} benchmarks, resulting in 464 unique programming problems. We create a 348/116 train/test split. To simulate realistic statistical heterogeneity, the 348 training problems are distributed among clients using a Dirichlet distribution with a concentration parameter $\alpha=0.1$, ensuring a non-IID data landscape. An independent set of 128 problems is held out as the \emph{knowledge-transfer set} ($\mathcal{X}_u$), used exclusively for knowledge distillation and unseen during local training.

\paragraph{Evaluation Metrics.} Our primary metric is \textbf{Pass@k} ($k \in \{1, 5, 10\}$)~\cite{chen2021evaluating}, which measures the functional correctness of the generated code against unit tests. To assess syntactic and structural quality, we also report \textbf{CodeBLEU}~\cite{ren2020codebleu}.

\paragraph{Client Fleet and Models.}
To probe heterogeneity (RQ3), our main experiments feature a federation of \textbf{6} clients. Each client hosts a distinct open-source LLM backbone from a pool of models in the $\approx$500M parameter class, including \texttt{Qwen1.5-0.5B}, \texttt{Cerebras-GPT-590M}, \texttt{bloom-560m}, and \texttt{pythia-410m}. This 6-client setup is a deliberate stress test for high model/data heterogeneity, where naive collaboration fails and orchestration is most critical. All models are fine-tuned using LoRA~\cite{Hu2021LoRA}, with ranks empirically optimized to keep trainable parameters between 2.2–3.0\% of the total, significantly reducing communication payloads (details in Table~\ref{tab:exp_client_config}).

\begin{table}[!ht]
    \centering
    \resizebox{\columnwidth}{!}{%
    \begin{tabular}{@{}llccc@{}}
        \toprule
        \textbf{Client} & \textbf{Backbone Model} & \textbf{\# Params} & \textbf{Train \%} & \textbf{Train/Val} \\
        \midrule
        C0 & \texttt{Qwen1.5-0.5B}           & 475M & 2.39\% & 45 / 12 \\
        C1 & \texttt{Cerebras-GPT-590M}      & 604M & 2.34\% & 43 / 11 \\
        C2 & \texttt{bloom-560m}             & 572M & 2.20\% & 44 / 12 \\
        C3 & \texttt{pythia-410m}            & 418M & 3.01\% & 46 / 12 \\
        C4 & \texttt{Qwen1.5-0.5B}           & 475M & 2.39\% & 54 / 14 \\
        C5 & \texttt{Cerebras-GPT-590M}      & 604M & 2.34\% & 44 / 11 \\
        \bottomrule
    \end{tabular}}
    \caption{Heterogeneous 6-client configuration for our primary \textsc{KNEXA-FL} experiments. All backbones are tuned with empirically-optimized LoRA settings.}
    \label{tab:exp_client_config}

\end{table}

\paragraph{Baselines.} We compare \textsc{KNEXA-FL} against a rigorous set of baselines:
\begin{itemize}
    \item \textbf{\texttt{LocalOnly}}: Establishes the performance frontier without collaboration. We report the average client performance after 12 rounds of isolated local fine-tuning.
    \item \textbf{\texttt{FedID-CentralKD}}: A modern centralized baseline, adapted from the Federated Interactive Distillation (\texttt{FedID}) framework~\cite{ma-etal-2023-fedid}. A server model (\texttt{Qwen-0.5B}) learns from client predictions on $\mathcal{X}_u$ via a confidence-weighted ensemble, then acts as a central teacher for all clients via text-level KD. This is privacy-preserving (server never sees private data).
    \item \textbf{\texttt{Central-KD}}: A strong centralized baseline inspired by FedMD~\cite{Li2019FedMD}. In each round, all clients submit their logits on the transfer set $\mathcal{X}_u$ to a central server. The server averages these logits to create an "ensemble teacher" distribution, which is then broadcast back to all clients for distillation. This represents a conventional approach to heterogeneous federated learning.
    \item \textbf{\texttt{Heuristic-P2P}}: A non-learning ablation using identical AKD. It replaces the CPM with a static heuristic: greedily pair clients to maximize data (JS) divergence, with the higher-performer as teacher.
    \item \textbf{\texttt{Random-P2P}}: A direct ablation of the CPM. This method uses the same P2P knowledge distillation as \textsc{KNEXA-FL}, but teacher-student pairings are selected uniformly at random in each of the 20 collaboration rounds.
\end{itemize}

\paragraph{Implementation Details.}
All experiments were conducted on a mix of NVIDIA H100s and A100s with clients simulated in parallel. Fixed random seed of 42 for reproducibility. Additional hyperparameters in the Appendix.

\subsection{Results and Analysis}
\label{subsec:exp_results}

\paragraph{RQ1: Overall Performance.}
    Table~\ref{tab:main_results} validates our claims. The \texttt{LocalOnly} baseline (2.22\% Pass@1) confirms collaboration is necessary. Our centralized baselines show extreme fragility under heterogeneity: the modern \texttt{FedID-CentralKD} failed to converge (1.11\% Pass@1), reinforcing the \texttt{Central-KD} baseline's collapse (peak 18.33\% $\to$ final 2.00\%) and confirming aggregation is unreliable in this regime. The P2P baselines are most critical. While \texttt{Random-P2P} (8.89\% Pass@1) confirms P2P's value, the \texttt{Heuristic-P2P} baseline, which maximizes data diversity (JS divergence), performed \textbf{worse} (6.67\% Pass@1). This strongly suggests that naive, non-learning heuristics can be detrimental. In sharp contrast, \textbf{\textsc{KNEXA-FL}} achieves the highest performance (13.33\% Pass@1), a \textbf{50\%} relative gain over \texttt{Random-P2P} and \textbf{100\%} over the failing heuristic. This demonstrates that our CPM's \textit{learned} policy, which balances diversity and compatibility (Fig. 2c), successfully navigates the trade-offs for stable, superior performance.

\begin{table}[!ht]
    \centering
    \resizebox{\columnwidth}{!}{%
    \begin{tabular}{@{}lcccc@{}}
        \toprule
        \textbf{Method} & \textbf{Pass@1 (\%)} & \textbf{Pass@5 (\%)} & \textbf{Pass@10 (\%)} & \textbf{CodeBLEU} \\
        \midrule
        {LocalOnly}    & 2.22 & 5.42 & 5.55 & 0.260 \\
        \midrule
        {FedID-CentralKD} & 1.11 & 5.56 & 5.56 & 0.181 \\
        {Central-KD}    & 2.00 (18.33)\textsuperscript{†} & 7.80 & 10.00 & 0.268 \\
        \midrule
        {Heuristic-P2P}\textsuperscript{‡} & 6.67 & 16.67 & 27.78 & 0.392 \\
        {Random-P2P}    & 8.89 & 22.40 & 27.80 & 0.239 \\
        \textbf{\textsc{KNEXA-FL}}    & \textbf{13.33} & \textbf{31.25} & \textbf{44.44} & \textbf{0.344} \\
        \bottomrule
    \end{tabular}}
    \textsuperscript{†}\scriptsize{\texttt{Central-KD} was volatile; peaked at 18.33\% (6-client) but collapsed to 2.00\% (4-client instability analysis).}
    \textsuperscript{‡}\scriptsize{\texttt{Heuristic-P2P} was evaluated for representative restricted rounds  and data on the same 6-client setup.}
    \caption{Average performance on the global test set. \texttt{LocalOnly} is evaluated after 12 rounds of isolated training. Collaborative methods are evaluated after 20 rounds, unless otherwise noted. Best final performance is in bold.}
    \label{tab:main_results}

\end{table}k

\paragraph{The Instability of Centralized Distillation.}
A deeper look at \texttt{Central-KD} shows fundamental instability. Although it briefly peaked at 18.33\% Pass@1 (6 clients), a controlled 4-client run with full logging confirmed a collapse to 2.00\% final Pass@1. Forcing heterogeneous models to distill from a single averaged ``ensemble teacher'' overwrites specialized knowledge, triggering catastrophic forgetting, which \textsc{KNEXA-FL}'s targeted, utility-driven P2P exchanges avoid.

\paragraph{RQ2: Matchmaking Efficacy.}
The performance gap between \texttt{Random-P2P} and \textsc{KNEXA-FL} directly points to the efficacy of the CPM. To isolate this effect, we measured the peak performance achieved by any student on the knowledge-transfer set during collaboration (Table~\ref{tab:kd_effectiveness}). \textsc{KNEXA-FL}'s CPM-guided pairings enabled a student to achieve \textbf{86.70\%} Pass@1 on this set, a staggering \textbf{2.6$\times$} improvement over the best pairing found by \texttt{Random-P2P}. This demonstrates that the CPM is not merely avoiding bad pairings but is actively discovering and exploiting highly synergistic knowledge transfers that random chance is unlikely to find.

\begin{table}[!ht]
    \centering
    \begin{tabular}{@{}lc@{}}
        \toprule
        \textbf{Pairing Strategy} & \textbf{Peak Student Pass@1} \\
        \midrule
        \texttt{Random-P2P} & 33.33\% \\
        \textbf{\textsc{KNEXA-FL} (CPM-Guided)}   & \textbf{86.70\%} \\
        \bottomrule
    \end{tabular}
    \caption{Peak student Pass@1 achieved on the 128-sample transfer set during collaboration. This metric isolates the quality of knowledge transfer.}
    \label{tab:kd_effectiveness}

\end{table}

\subsubsection{Profiler Ablation: LinUCB under a Controlled Synthetic Regime}
\label{subsubsec:linucb_ablation}
The gains in Table~\ref{tab:kd_effectiveness} imply that \emph{who} collaborates with whom is decisive. To attribute this effect precisely to the LinUCB-driven Central Profiler/Matchmaker and to test its scalability beyond the six-client, real-model setup, we conducted a controlled ablation in a large, \textbf{synthetic} environment, an established protocol for bandit and FL research~\cite{lattimore2020bandit}.

The results, presented in Figure~\ref{fig:synthetic_ablation}, are definitive:
\begin{itemize}
    \item \textbf{Substantial and Scalable Gains:} The LinUCB-enhanced CPM delivers major performance improvements over random pairing, peaking at a \textbf{48.5\%} relative gain in the 32-client, high-heterogeneity case (Panel a). This advantage scales robustly, with performance consistently approaching the oracle upper bound as the federation grows (Panel b).
    \item \textbf{Learned Compatibility Trade-off:} The CPM learns a non-trivial strategy for exploiting diversity. While a naive `Hetero-Greedy` baseline maximizes Jensen-Shannon divergence, our CPM intelligently trades a small amount of diversity for a large gain in synergistic compatibility, explaining its superior performance (Panel c). This is achieved with highly practical computational efficiency (Panel d).
\end{itemize}
\noindent\textbf{Take-away.} Under this isolated, reproducible setting, the LinUCB-driven CPM demonstrably \emph{learns} a superior, scalable matchmaking policy that materially explains the end-to-end performance gains of \textsc{KNEXA-FL}, fully answering \textbf{RQ2}.

\paragraph{RQ3: Heterogeneous Federation.}
Per-client analysis of the 6-agent \textsc{KNEXA-FL} run shows that gains are distributed across the diverse federation. For instance, Client C2 (\texttt{bloom-560m}) evolved from a mid-tier performer to the strongest individual model, achieving a final local Pass@1 of 36.67\%. Even the smallest model, C3 (\texttt{pythia-410m}), significantly surpassed its isolated performance, demonstrating that profiler-guided collaboration effectively lifts the entire ecosystem, not just the strongest members.

\subsection{Discussion and Limitations}
\label{subsec:exp_discussion}

Our results show \textsc{KNEXA-FL} delivers substantial, stable gains with low overhead. The improvements stem from the CPM's learned P2P orchestration, contrasting sharply with the volatile, "one-size-fits-all" centralized baselines. By targeting knowledge sharing, \textsc{KNEXA-FL} mitigates the catastrophic forgetting seen in simpler schemes.

\paragraph{Limitations.}
Our study has limitations. Future work should: (i) validate on larger federations with realistic WAN latencies; (ii) explore more semantic (e.g., user-profile-based) data splits beyond Dirichlet partitioning; and (iii) benchmark against a wider array of advanced centralized FL optimizers.

Despite these constraints, the consistent and stable performance edge over strong baselines firmly positions profiler-guided P2P learning as a robust and promising paradigm for decentralized, collaborative LLM intelligence.

\section{Conclusion}
\label{sec:conclusion}

We introduced \textsc{KNEXA-FL}, a framework for orchestrated decentralization that resolves the trade-off between insecure centralized FL and inefficient random P2P collaboration. Its non-aggregating Central Profiler/Matchmaker (CPM) formulates P2P matchmaking as a contextual bandit problem, learning to optimize the collaborative graph. Empirically, on a heterogeneous code-generation task, our approach yields substantial gains ($\approx$50\% relative Pass@1 improvement over random P2P) and, critically, achieves stable convergence where a strong centralized distillation baseline catastrophically fails. Our work establishes learning-based orchestration as a core principle for robust decentralized AI. Future work includes: (i) scaling to larger federations, (ii) exploring more expressive (e.g., neural) bandit models, and (iii) integrating safeguards such as differential privacy, zero-knowledge proofs, and token-efficient disparity audits like TFDP~\cite{singh2025tfdp}.

\bibliography{aaai2026}

\appendix

\noindent\fbox{%
\parbox{0.97\linewidth}{%
\textbf{Code Availability:} The complete source code for the \textsc{KNEXA-FL} framework, all experiments, and analysis scripts is publicly available at: https://github.com/FujitsuResearch/knexa-fl
}%
}
\vspace{5mm}

\section{Detailed Algorithms}
\label{sec:appendix_algorithms}

This section provides the precise algorithmic specifications for the Central Profiler/Matchmaker (CPM) and the multi-phase asynchronous training loop of \textsc{KNEXA-FL}. The presentation follows the notation introduced in the main paper.

\subsection{CPM Matchmaking Logic}
\label{subsec:appendix_matchmaking}
Algorithm~\ref{alg:appendix_profiler_matchmaking} details the contextual combinatorial bandit procedure used by the CPM to select up to $K_p$ disjoint peer-to-peer (P2P) interactions per round. The CPM evaluates every admissible tuple $(a_i, a_j, K_{\text{rec}}, R_{ij})$, comprising a sender $a_i$, a receiver $a_j$, the recommended knowledge-exchange protocol $K_{\text{rec}}$ (e.g., Adaptive Knowledge Distillation, AKD), and a role assignment $R_{ij}$, with an upper-confidence-bound (UCB) score. Pairs are greedily selected until either (i) the pair budget $K_p$ is reached or (ii) no agents remain unmatched.

\begin{algorithm}[!ht]
\caption{CPM: LinUCB-based Matchmaking}
\label{alg:appendix_profiler_matchmaking}
\begin{algorithmic}[1]
\Require \textbf{Profiles} $\{\mathbf{p}_i^{(t)}\}_{i=1}^N$, \textbf{LinUCB state} $(\mathbf{A}, \mathbf{b})$, \textbf{exploration} $\beta$, \textbf{pair budget} $K_p$
\State $\hat{\bm{\theta}} \leftarrow \mathbf{A}^{-1}\mathbf{b}$ \Comment{Current estimate of reward model weights}
\State $\mathcal{C} \leftarrow \emptyset$, $\mathcal{E}_t \leftarrow \emptyset$, $\mathcal{U} \leftarrow \{a_1, \dots, a_N\}$
\ForAll{unordered pairs $(a_i, a_j)$ with $a_i, a_j \in \mathcal{U}$}
    \ForAll{admissible $(K_{\text{rec}}, R_{ij})$}
        \State $\mathbf{x} \leftarrow \varphi(\mathbf{p}_i^{(t)}, \mathbf{p}_j^{(t)}, K_{\text{rec}}, R_{ij})$ \Comment{Construct context vector}
        \State $u \leftarrow \hat{\bm{\theta}}^{\top}\mathbf{x} + \beta \sqrt{\mathbf{x}^{\top}\mathbf{A}^{-1}\mathbf{x}}$ \Comment{Calculate UCB score}
        \State Append $(a_i, a_j, K_{\text{rec}}, R_{ij}, u)$ to $\mathcal{C}$
    \EndFor
\EndFor
\State Sort $\mathcal{C}$ in descending order by UCB score $u$
\ForAll{$(a_i, a_j, K_{\text{rec}}, R_{ij}, u) \in \mathcal{C}$ \textbf{in order}}
    \If{$|\mathcal{E}_t| < K_p$ \textbf{and} $a_i, a_j \in \mathcal{U}$}
        \State Add $(a_i, a_j, K_{\text{rec}}, R_{ij})$ to $\mathcal{E}_t$
        \State $\mathcal{U} \leftarrow \mathcal{U} \setminus \{a_i, a_j\}$ \Comment{Remove agents from available pool}
    \EndIf
\EndFor
\State \Return $\mathcal{E}_t$
\end{algorithmic}
\end{algorithm}

\subsection{End-to-End Asynchronous Training Round}
\label{subsec:appendix_main_loop}
Algorithm~\ref{alg:appendix_main_loop} expands the high-level protocol (Algorithm 1 in the main paper) into an explicit four-phase sequence executed at every communication round $t \in \{1, \dots, T\}$. All network communication is authenticated and encrypted (e.g., via mTLS) to guarantee the confidentiality and integrity of the exchanged profiles and knowledge packages.

\begin{algorithm}[!ht]
\caption{\textsc{KNEXA-FL}: Asynchronous Training Round $t$}
\label{alg:appendix_main_loop}
\begin{algorithmic}[1]
\Statex \textbf{Inputs:} Agent set $\mathcal{A}=\{a_1,\dots,a_N\}$, CPM $\mathcal{P}$, public transfer set $\mathcal{X}_u$
\vspace{2pt}

\Statex \rule{\linewidth}{0.4pt}
\Statex \textit{Phase 1 (Agent-side local updates, parallel):}
\ForAll{$a_i \in \mathcal{A}$ \textbf{in parallel}}
    \State Perform $E$ local epochs on private data $D_i$: $\bm{\phi}_i \leftarrow \bm{\phi}_i - \eta\nabla_{\bm{\phi}_i}\mathcal{L}_i$.
    \State Prepare knowledge package $\kappa_i$ (e.g., logits on $\mathcal{X}_u$) and profile $\mathbf{p}_i^{(t)}$.
    \State Securely transmit profile $\mathbf{p}_i^{(t)}$ to $\mathcal{P}$.
\EndFor

\Statex \rule{\linewidth}{0.4pt}
\Statex \textit{Phase 2 (CPM-side matchmaking, periodic):}
\State On quorum or timer: obtain latest profiles $\{\mathbf{p}_i^{(t)}\}$; compute $\mathcal{E}_t$ via Alg.~\ref{alg:appendix_profiler_matchmaking}.
\State Dispatch pairing directives to the involved agents.

\Statex \rule{\linewidth}{0.4pt}
\Statex \textit{Phase 3 (P2P knowledge exchange, parallel):}
\ForAll{$(a_s, a_r, K_{\text{rec}}, R_{sr}) \in \mathcal{E}_t$ \textbf{in parallel}}
    \State Establish secure, ephemeral channel; $a_s$ transmits $\kappa_s$ to $a_r$.
    \State $a_r$ integrates $\kappa_s$ using protocol $K_{\text{rec}}$ (AKD by default).
\EndFor

\Statex \rule{\linewidth}{0.4pt}
\Statex \textit{Phase 4 (Reward reporting \& CPM update, asynchronous):}
\ForAll{receiving agents $a_r$}
    \State Compute reward $r^{(t)}_{sr}$ (Eq. 3 in main text) and send $(\mathbf{x}_{sr}^{(t)}, r^{(t)}_{sr})$ to $\mathcal{P}$.
    \State \textit{On receipt, $\mathcal{P}$ updates bandit model:} $\mathbf{A} \leftarrow \mathbf{A} + \mathbf{x}\mathbf{x}^{\top}$, $\mathbf{b} \leftarrow \mathbf{b} + r\mathbf{x}$.
\EndFor
\end{algorithmic}
\end{algorithm}

\section{Experimental Setup Details}
\label{sec:appendix_setup}

All experiments were conducted under a fixed random seed (\texttt{42}) to ensure reproducibility of data partitioning, model initialization, and matchmaking. The software stack included \texttt{PyTorch 2.3}, \texttt{Hugging Face Transformers 4.43}, and \texttt{Accelerate 0.29}. Training was performed using \texttt{FP16} precision with automatic mixed-precision scaling.

\subsection{Core Hyperparameters}
\label{subsec:appendix_hparams}
Table~\ref{tab:appendix_hyperparams} summarizes the key hyperparameters shared across all real-model experiments.

\begin{table*}[!ht]
\centering
\resizebox{0.6\textwidth}{!}{%
\begin{tabular}{@{}ll@{}}
\toprule
\textbf{Category} & \textbf{Value / Specification} \\
\midrule
Optimizer & AdamW (\texttt{betas=(0.9, 0.98)}, \texttt{eps=1e-6}) \\
Learning Rate & $3 \times 10^{-5}$ with linear warmup (2\% of steps) \\
Local Batch Size & 8 (gradient accumulated to an effective size of 32) \\
Local Epochs / Round & 1 \\
Comm. Rounds ($T$) & 20 (for main experiments), 100 (for synthetic) \\
\midrule
PEFT Method & LoRA, rank tuned per model (2.2–3.0\% trainable) \\
KD Temperature $\mathcal{T}$ & 1.5–2.5 (linearly annealed over training) \\
KD Weight $\alpha_{\textsc{kd}}$ & 0.2 (initial) $\rightarrow$ 0.5 (final), linear schedule \\
\midrule
Compute Hardware & NVIDIA A100 80GB or H100 90GB (mixed fleet) \\
Parallelism & One client per GPU; CPM on CPU (\texttt{Intel Xeon 6338}) \\
\bottomrule
\end{tabular}}
\caption{Global hyperparameter configuration for all experiments.}
\label{tab:appendix_hyperparams}
\end{table*}

\subsection{Six-Client Heterogeneous Federation}
\label{subsec:appendix_fed_config}
The primary empirical study (Sections 4-5 in the main paper) employs a federation of six heterogeneous clients. Table~\ref{tab:appendix_client_config} lists the backbone models, parameter counts, and data partition sizes, which were generated using a Dirichlet distribution ($\alpha=0.1$) over the combined \texttt{HumanEval+MBPP} training set.

\begin{table}[t]
\centering
\resizebox{\columnwidth}{!}{%
\begin{tabular}{@{}llccc@{}}
\toprule
\textbf{Client} & \textbf{Backbone Model} & \textbf{Total Params} & \textbf{Trainable \%} & \textbf{Train / Val} \\
\midrule
C0 & \texttt{Qwen1.5-0.5B} & 475M & 2.39\% & 45 / 12 \\
C1 & \texttt{Cerebras-GPT-590M} & 604M & 2.34\% & 43 / 11 \\
C2 & \texttt{BLOOM-560M} & 572M & 2.20\% & 44 / 12 \\
C3 & \texttt{Pythia-410M} & 418M & 3.01\% & 46 / 12 \\
C4 & \texttt{Qwen1.5-0.5B} & 475M & 2.39\% & 54 / 14 \\
C5 & \texttt{Cerebras-GPT-590M} & 604M & 2.34\% & 44 / 11 \\
\bottomrule
\end{tabular}}
\caption{Client fleet configuration for real-model experiments. \emph{Trainable \%} denotes the proportion of parameters updated via LoRA.}
\label{tab:appendix_client_config}
\end{table}

\section{LinUCB CPM Simulation Details}
\label{sec:appendix_cpm}

To rigorously stress-test the contextual-bandit CPM in isolation, we executed large-scale synthetic simulations (up to 64 clients) following the design protocol in \citet{lattimore2020bandit}. These experiments allow for a clean analysis of the learning algorithm's behavior, free from the confounding variables of real-world model training.

\begin{figure*}[!ht]
    \centering
    \includegraphics[width=0.85\linewidth]{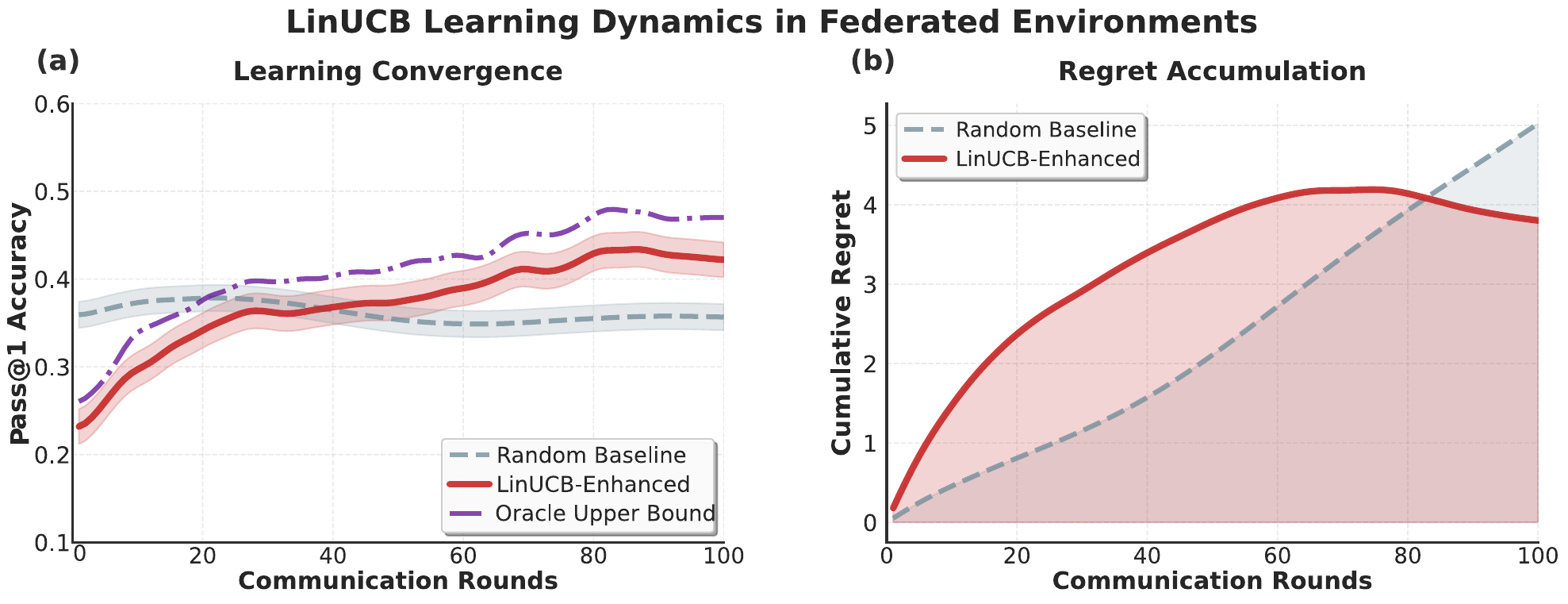}
    \caption{LinUCB-based CPM learning dynamics over 100 communication rounds in synthetic federated environments.
    (a) Learning Convergence: Pass@1 accuracy trajectories demonstrate the LinUCB algorithm's ability to learn optimal pairing strategies over time. The LinUCB-enhanced approach (red) shows steady improvement and convergence toward the oracle upper bound (purple dashed), while the random baseline (gray dashed) plateaus at suboptimal performance. The shaded regions represent 95\% confidence intervals.
    (b) Regret Accumulation: Cumulative regret quantifies the performance gap between each method and the oracle policy. LinUCB's regret stabilizes after approximately 60 rounds, indicating convergence to a near-optimal policy, while the random baseline's regret grows linearly. The plateauing of LinUCB's regret curve confirms the algorithm's successful exploitation of learned pairing patterns.
    These synthetic experiments validate the theoretical foundations of our contextual bandit approach and demonstrate why intelligent matchmaking is crucial for maximizing collaborative gains in heterogeneous federated learning environments.}
    \label{fig:appendix_learning_curves}
\end{figure*}

Figure~\ref{fig:appendix_learning_curves} provides a detailed visualization of the learning dynamics, complementing the summary results presented in Figure 3 of the main manuscript. The plots for learning convergence (a) and cumulative regret (b) offer strong visual evidence for the efficacy of the LinUCB algorithm. They confirm that the CPM learns a near-optimal matchmaking policy over time, leading to monotonic performance improvements and outperforming the random baseline by a significant margin. The sub-linear regret curve for KNEXA-FL is characteristic of an efficient learning algorithm, contrasting sharply with the linear regret of the non-adaptive random strategy.

\section{Detailed Experimental Results}
\label{sec:appendix_results}

This section provides a more granular analysis of the experimental results, including per-client performance trajectories and a principled discussion of the baseline methods' failure modes.

\subsection{Aggregate Performance Summary}
\label{subsec:appendix_summary}
Table~\ref{tab:appendix_summary} consolidates the final global test performance after 20 communication rounds for all collaborative methods and after 12 isolated rounds for the \texttt{LocalOnly} baseline.

\begin{table}[!ht]
\centering
\resizebox{\columnwidth}{!}{%
\begin{tabular}{@{}lcccc@{}}
\toprule
\textbf{Method} & \textbf{Pass@1 (\%)} & \textbf{Pass@5 (\%)} & \textbf{Pass@10 (\%)} & \textbf{CodeBLEU} \\
\midrule
{LocalOnly}    & 2.22 & 5.42 & 5.55 & 0.260 \\
\midrule
{FedID-CentralKD} & 1.11 & 5.56 & 5.56 & 0.181 \\
{Central-KD}\textsuperscript{†}    & 2.00 & 7.80 & 10.00 & 0.268 \\
\midrule
{Heuristic-P2P}\textsuperscript{‡} & 6.67 & 16.67 & 27.78 & 0.392 \\
{Random-P2P}    & 8.89 & 22.40 & 27.80 & 0.239 \\
\textbf{\textsc{KNEXA-FL}}    & \textbf{13.33} & \textbf{31.25} & \textbf{44.44} & \textbf{0.344} \\
\bottomrule
\end{tabular}}
\vspace{4pt}
\scriptsize{\textsuperscript{†}\texttt{Central-KD} was volatile; peaked at 18.33\% (6-client) but collapsed to 2.00\% (4-client instability analysis).}
\scriptsize{\textsuperscript{‡}\texttt{Heuristic-P2P} was evaluated for restricted rounds and on data due to time/compute constraints, to ablate learning vs. a static heuristic.}
\caption{Final average performance on the 116-problem global test set. Best global-test performance is in bold. Footnotes match the main paper.}
\label{tab:appendix_summary}
\end{table}

\subsection{Baseline Instability and Inefficiency Analysis}
\label{subsec:appendix_collapse}

\paragraph{Central-KD: Catastrophic Collapse.} The \texttt{Central-KD} baseline, which forces all clients to distill from a single, globally-averaged teacher distribution, exhibited severe training instability. This phenomenon arises from an inherent incompatibility between global logit averaging and a highly heterogeneous federation. Forcing diverse, specialized PEFT-based models to conform to a single ensemble teacher leads to catastrophic forgetting, as locally acquired knowledge is overwritten by destructive updates. Table~\ref{tab:appendix_central_kd_collapse} quantifies this collapse by showing the dramatic drop from peak to final performance, confirming that this approach is fundamentally unreliable in heterogeneous settings.

\begin{table}[!ht]
\centering
\resizebox{\columnwidth}{!}{%
\begin{tabular}{@{}lccc@{}}
\toprule
\textbf{Client (Model)} & \textbf{Peak Pass@1} & \textbf{Final Pass@1} & \textbf{Degradation} \\
\midrule
C0 (Qwen-0.5B) & 13.33\% & 0.00\% & -13.33 pp \\
C1 (Cerebras-590M) & 0.00\% & 0.00\% & 0.00 pp \\
C2 (BLOOM-560M) & 20.00\% & 0.00\% & -20.00 pp \\
C3 (Pythia-410M) & 40.00\% & 8.00\% & -32.00 pp \\
\midrule
\textbf{Average} & \textbf{18.33\%} & \textbf{2.00\%} & {\textbf{-16.33 pp}} \\
\bottomrule
\end{tabular}}
\caption{Performance collapse of the \texttt{Central-KD} baseline, comparing peak Pass@1 (\%) to final Pass@1 after 14 rounds on the global test set.}
\label{tab:appendix_central_kd_collapse}
\end{table}

\paragraph{Random-P2P: Inefficiency of Unguided Collaboration.} The \texttt{Random-P2P} baseline avoids catastrophic collapse but demonstrates the statistical inefficiency of unguided collaboration. Its final Pass@1 of 8.89\% is substantially lower than the 13.33\% from \textsc{KNEXA-FL}. This gap highlights the value of the CPM: random pairings are unlikely to consistently identify the most synergistic knowledge transfers, leading to slower convergence and a lower overall performance ceiling. The CPM's learned policy is the decisive factor that elevates \textsc{KNEXA-FL} beyond both the instability of centralization and the inefficiency of randomness.

\subsection{Per-Client Improvements under KNEXA-FL}
\label{subsec:appendix_perclient}
Table~\ref{tab:appendix_perclient} reports the individual Pass@1 improvements for each client, comparing their final \textsc{KNEXA-FL} performance to their \texttt{LocalOnly} starting point. The results show that all six clients benefit significantly, with the most dramatic gains seen by Client C2. This confirms that CPM-guided collaboration effectively raises the performance of the entire federation, rather than merely amplifying the strongest members.

\begin{table}[!ht]
\centering
\resizebox{\columnwidth}{!}{%
\begin{tabular}{@{}lccc@{}}
\toprule
\textbf{Client} & \textbf{LocalOnly (\%)} & \textbf{KNEXA-FL (\%)} & \textbf{Improvement ($\Delta$)} \\
\midrule
C0 (Qwen) & 2.22 & 11.11 & \textbf{+8.89 pp} \\
C1 (Cerebras) & 0.00 & 6.67 & \textbf{+6.67 pp} \\
C2 (BLOOM) & 6.67 & 36.67 & \textbf{+30.00 pp} \\
C3 (Pythia) & 0.00 & 8.89 & \textbf{+8.89 pp} \\
C4 (Qwen) & 2.22 & 10.00 & \textbf{+7.78 pp} \\
C5 (Cerebras) & 0.00 & 7.78 & \textbf{+7.78 pp} \\
\bottomrule
\end{tabular}}
\caption{Per-client Pass@1 (\%) on the global test set. Absolute improvement ($\Delta$) is relative to the \texttt{LocalOnly} baseline.}
\label{tab:appendix_perclient}
\end{table}

\section{Expanded Reproducibility Checklist}
\label{sec:appendix_checklist}

To ensure full transparency and replicability, this supplementary material provides the following artifacts:
\begin{itemize}
    \item \textbf{Detailed Algorithms:} Complete pseudocode for both the CPM matchmaking logic and the end-to-end asynchronous training round are provided in Algorithms~\ref{alg:appendix_profiler_matchmaking} and \ref{alg:appendix_main_loop}.
    \item \textbf{Comprehensive Hyperparameters:} The full training configuration, including optimizer settings, learning rate schedules, and hardware specifications, is detailed in Table~\ref{tab:appendix_hyperparams}.
    \item \textbf{Federation Configuration:} The precise specification of the heterogeneous client fleet, including backbone models and data partitions, is provided in Table~\ref{tab:appendix_client_config}.
    \item \textbf{Granular Results:} Detailed tables showing per-client performance gains (Table~\ref{tab:appendix_perclient}) and the quantitative collapse of the centralized baseline (Table~\ref{tab:appendix_central_kd_collapse}) are included.
    \item \textbf{Public Code Release:} All source code, experiment scripts, configuration files, and scripts to reproduce the synthetic benchmarks are publicly available at {https://github.com/FujitsuResearch/knexa-fl}.
\end{itemize}

\begin{algorithm}[t]
\caption{Heuristic-P2P (Hetero-Greedy) Pairing}
\label{alg:hetero_greedy}
\begin{algorithmic}[1]
\Require Client set $\mathcal{C}=\{a_1, \dots, a_N\}$, per-client data distributions $\{d_i\}$, per-client recent performance $\{u_i\}$, pair budget $K_p = \lfloor N/2 \rfloor$.
\State Initialize candidate list $\mathcal{S} \leftarrow \emptyset$, matched pairs $\mathcal{E}_t \leftarrow \emptyset$, and set of available agents $\mathcal{U} \leftarrow \mathcal{C}$.
\ForAll{unordered pairs $(a_i, a_j)$ with $a_i, a_j \in \mathcal{U}$}
    \State Compute score $s_{ij} = \text{JS}(d_i, d_j)$.
    \State Add $(a_i, a_j, s_{ij})$ to $\mathcal{S}$.
\EndFor
\State Sort $\mathcal{S}$ by score $s_{ij}$ in descending order.
\ForAll{sorted tuple $(a_i, a_j, s_{ij})$ in $\mathcal{S}$}
    \If{$|\mathcal{E}_t| < K_p$ \textbf{and} $a_i \in \mathcal{U}$ \textbf{and} $a_j \in \mathcal{U}$}
        \State Assign roles: $a_t \leftarrow \arg\max(u_i, u_j)$, $a_s \leftarrow \arg\min(u_i, u_j)$.
        \State Add directed pair $(a_s, a_t)$ to $\mathcal{E}_t$.
        \State $\mathcal{U} \leftarrow \mathcal{U} \setminus \{a_i, a_j\}$.
    \EndIf
\EndFor
\State \Return $\mathcal{E}_t$.
\end{algorithmic}
\end{algorithm}

\section{Rebuttal Additions and Protocol Details}
\label{sec:appendix_rebuttal}

To address reviewer feedback and ensure full reproducibility, this section provides explicit details on the CPM's inputs, the disjoint-pair matching algorithm, and the implementation of our new baselines.

\subsection{CPM Context Vector and Reward Normalization}
\label{subsec:appendix_cpm_details}

As promised in our rebuttal, we specify the construction of the CPM's inputs and reward signal, as referenced in Section 3.3 of the main paper.

\paragraph{Context Vector Construction.}
The abstract agent profile $\mathbf{p}_i \in \mathbb{R}^{d_p}$ submitted by agent $a_i$ is a concatenation of features. For our experiments, $d_p=32$. The profile includes:
\begin{itemize}
    \item \textbf{Static Features (8 dims):} One-hot encodings of the agent's LLM backbone family (e.g., Qwen, Pythia, BLOOM).
    \item \textbf{Dynamic Performance Features (12 dims):} Recent local validation metrics (e.g., Pass@1, CodeBLEU, perplexity) and their deltas from the previous round.
    \item \textbf{Data Distribution Features (12 dims):} A normalized histogram representing the agent's local data distribution over $k=12$ problem categories (e.g., string manipulation, algorithms, data structures).
\end{itemize}
When the CPM considers a pair $(a_i, a_j)$, the final context vector $\mathbf{x}_{ij} \in \mathbb{R}^{2d_p}$ is constructed by concatenating the individual profiles $\mathbf{x}_{ij} = [\mathbf{p}_i, \mathbf{p}_j]$. This simple concatenation allows the LinUCB model's linear weights $\hat{\bm{\theta}}$ to learn the cross-agent feature interactions.

\paragraph{Reward Normalization.}
The raw reward signal $r_{ij}^{(t)}$ from Eq. 3 (main paper) can be volatile. To stabilize the LinUCB algorithm, we apply running z-score normalization. The CPM maintains a running mean $\mu_r$ and standard deviation $\sigma_r$ of all rewards it has observed. The reward $r_{ij}^{(t)}$ reported by agent $a_i$ is normalized before being used to update the bandit model:
$$
r_{\text{norm}}^{(t)} = \frac{r_{ij}^{(t)} - \mu_r}{\sigma_r + \epsilon}
$$
where $\epsilon=10^{-6}$ is a small constant for numerical stability. This ensures the reward signal has zero mean and unit variance, preventing large reward values from disproportionately skewing the bandit's parameter updates.

\subsection{Baseline Algorithm Specifications}
\label{subsec:appendix_baseline_algos}

We provide the algorithmic details for the new baselines added for the camera-ready version.

\paragraph{FedID-CentralKD.}
This baseline, adapted from \texttt{FedID}~\cite{ma-etal-2023-fedid}, was implemented as described in the main paper's `Baselines` paragraph (Section 4.1). A central \texttt{Qwen-0.5B + LoRA} model was trained for 20 rounds using text-level KD. The ensemble teacher signal was created by collecting decoded text from all 6 clients on public prompts, then performing confidence-weighted majority voting over normalized code sequences. The central model's validation loss on a private held-out set steadily decreased, but as shown in Table 4 in the main paper, this central knowledge failed to transfer effectively to the heterogeneous clients, resulting in 1.11\% Pass@1.

\paragraph{Heuristic-P2P (Hetero-Greedy).}
This baseline replaces the CPM's learned policy (Algorithm 2) with a static, non-learning heuristic. The algorithm, promised in our rebuttal, is specified in Algorithm~\ref{alg:hetero_greedy}. It greedily selects disjoint pairs that maximize JS divergence of their data distributions. As reported in Table 4, this approach (6.67\% Pass@1) was detrimental, performing worse than random pairings (8.89\% Pass@1).

\end{document}